\documentclass{article}

\usepackage{arxiv}

\usepackage[utf8]{inputenc} 
\usepackage[T1]{fontenc}    
\usepackage{hyperref}       
\usepackage{url}            
\usepackage{booktabs}       
\usepackage{amsfonts}       
\usepackage{amsmath}       
\usepackage{nicefrac}       
\usepackage{microtype}      
\usepackage{lipsum}
\usepackage{graphicx}
\usepackage[dvipsnames]{xcolor}

\usepackage{etoc}
\usepackage{enumitem}
\usepackage{float}

\graphicspath{ {./figs/} }

\usepackage{tcolorbox}
\usepackage{multicol}
\tcbuselibrary{skins,breakable}

\tcbset{
	breakable, nobeforeafter,
}

\newtcolorbox[auto counter,number freestyle={\noexpand Prompt \noexpand\arabic{\tcbcounter}}]{prompt}[2][]{
	label = #1,
	title={\thetcbcounter \quad  #2},
}

\newtcolorbox[auto counter,number freestyle={\noexpand Example \noexpand\arabic{\tcbcounter}}]{example}[2][]{
	label=#1,
	title={\thetcbcounter \quad   #2},
	coltitle = black,
	colbacktitle=gray!20!white,
	colback=gray!1!white
}

\definecolor{codebg}{gray}{0.98}
\definecolor{codeframe}{gray}{0.8}
\newcommand{\code}[1]{\fcolorbox{codeframe}{codebg}{\color{black}\!\texttt{#1}\!}}

\usepackage{titlesec}
\titlespacing*{\subsubsection}
{0pt}{.1ex plus .1ex minus .1ex}{.1ex plus .1ex}

\title{Information Theory of Meaningful Communication}

\author{
	\href{https://orcid.org/0000-0002-1422-5505}{\includegraphics[scale=0.06]{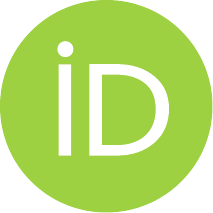}\hspace{1mm}Doron Sivan} \\
	Department of Brain Sciences\\
	Weizmann Institute of Science\\
	Rehovot 76100, Israel \\
	\texttt{doron.sivan@weizmann.ac.il} \\
	\And
	\href{https://orcid.org/0000-0002-5661-4349}{\includegraphics[scale=0.06]{orcid.pdf}\hspace{1mm}Misha Tsodyks} \\
	Department of Brain Sciences\\
	Weizmann Institute of Science\\
	Rehovot, 76100 Israel \\
	School of Natural Sciences\\
	Institute for Advanced Study\\
	Princeton, NJ \\
	\texttt{misha@weizmann.ac.il} \\
}

\begin{document}
\maketitle
\begin{abstract}
	In Shannon's seminal paper, entropy of printed English, treated as a stationary stochastic process, was estimated to be roughly 1 bit per character. However, considered as a means of communication, language differs considerably from its printed form: (i) the units of information are not characters or even words but clauses, i.e. shortest meaningful parts of speech; and (ii) what is transmitted is principally the meaning of what is being said or written, while the precise phrasing that was used to communicate the meaning is typically ignored. In this study, we show that one can leverage recently developed large language models to quantify information communicated in meaningful narratives in terms of bits of meaning per clause. 
\end{abstract}


\section{Introduction}
Natural language typically occurs during dialogues, where people talk (or write) to each other by exchanging 'messages', which are understood in the context of the previous conversation. 
These messages could be very short, like 'yes' or 'really?', or long, like e.g. somebody recollecting a recent trip to you.
In this study, we focus on the latter case by analyzing personal experience spoken narratives collected by Labov for his pioneering studies of narrative structures \cite{Labov1966}. 
In collecting these narratives, Labov asked his subjects a few introductory questions, like 'Was there any time when your life was in danger?', and when answered 'Yes', he further asked, 'What happened?'. 
The set of 20 narratives analyzed for our study span a large range of lengths, from 12 to 191 clauses. 
They were annotated and segmented into clauses by Labov \cite{Labov2013}, and we are using these segmented annotations for our analysis\footnote{
	The segmentation is for each independent clause and its dependent clauses \cite{Labov2013}. 
	For simplicity, we refer to it in this paper as a "clause" (see the Supplementary Information, section \ref{si:narratives}, for elaboration).}.

The goal of the study is to develop information theory of language as a means of communication, by quantifying information communicated by the narrator with each clause. 
Unlike the case of printed English considered by Shannon \cite{Shannon1951}, we are interested in the information contained in the \textit{meaning} of each clause rather than in the exact phrasing used (see e.g. \cite{Gunduz2022}). The rationale for this approach is that when exposed to meaningful language, people typically focus on the meaning of what is communicated to them, at the expense of the exact wording that is quickly forgotten, see e.g. \cite{Sachs1967}. 

\section{Information theory of meaningful communicative English}
\label{sec:headings}
We first develop our formulation in terms of exact phrasing of narratives and get back to the issue of meaning later. 
We consider a narrative $N$ as a finite sequence of clauses: 
\begin{equation}
	N = {C_1, C_2, ..., C_L} 
\end{equation}
where $C_i$ stands for the clause with position $i$, and $L$ is the narrative length in terms of the number of clauses. 
We can define $P(N|c)$ as the probability of sampling a narrative $N$ from the ensemble of narratives given a particular context, $c$ (e.g., spoken narratives that describe life-threatening events of narrators coming from a particular socioeconomic group).
This probability is used to define the information content of the narrative:
\begin{align}
	I_N = -\log P(N|c) \,\,.
	\label{info_nar}
\end{align}
Alternatively, $I_N$ is the smallest number of bits needed to encode it (in an optimal encoding of the ensemble of narratives) \cite{Shannon1948}.

Since the probability of a narrative is given by the product of conditional probabilities for all clauses, one can also define the information in each clause as
\begin{align}
	I_i &= -\log P(C_i|c_{<i}) \,\,, \label{eq:info_clause}
\end{align}
where $c_{<i}$ includes $c$ and the part of the narrative that precedes the clause $C_i$, so that $I_N = \sum_{i=1}^L I_i$.

In his pioneering work, Shannon estimated the entropy of printed English using a human subject who was trying to guess the next character given the previous ones \cite{Shannon1951}, with the result of about 1 bit per character (bpc). 
This result was corroborated by similar, but more controlled, experiments, e.g. \cite{Cover1978}.
Albeit considerable efforts, until the turn of the century, language models and compression methods were still behind at about 2 bpc \cite{Brown1992,Schurmann1996,Mahoney1999}.
Only recently, with the rise of Transformers and large language models (LLMs), 1 bpc was achieved \cite{Dai2019,Radford2019} (along with impressive results in many other natural language tasks, e.g. \cite{Brown2020}). 
Thus, in this study, we use LLMs to estimate $P(C_i|c_{<i})$, as they are specifically trained to approximate the conditional probabilities of each new token (usually word) given the context.
In particular, we use Llama3.1-70B-Instruct open-source LLM \cite{Llama3_2024}. 

\begin{figure}[htb!]
	\centering
	\includegraphics[width=\textwidth]{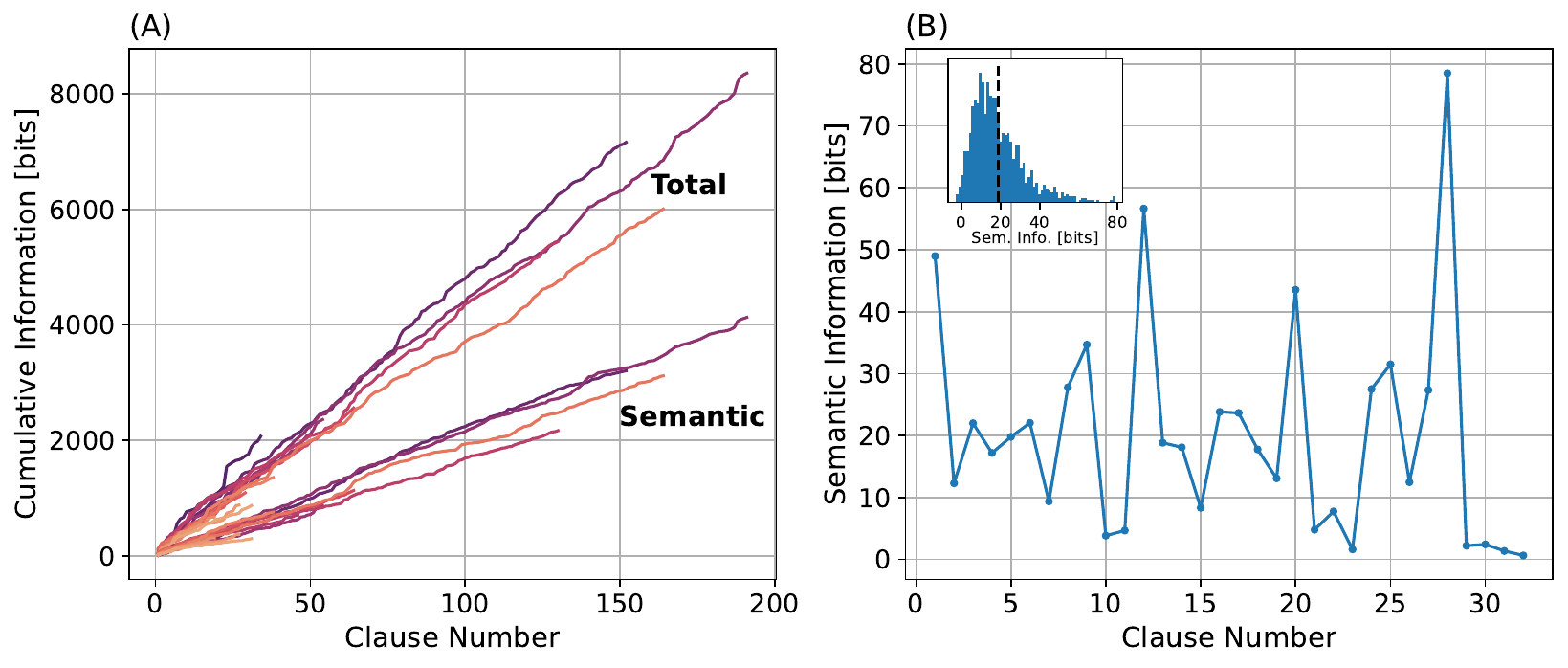}
	\caption{ {\it Semantic information in Labov narratives}. 
		(A) Cumulative total and semantic information ($\sum_{j=1}^{i} I_j$ and $\sum_{j=1}^{i} I_{M,j}$, respectively) up to clause $C_i$, as a function of clause number, $i$, for 20 Labov narratives (each narrative is represented by a unique color, with matching colors corresponding to the same narrative).
		The average slope of these curves is approximately 40 and 20 bits/clause, for total and semantic information, respectively.
		Thus, about half of the information contained in each clause is due to its meaning, while the other half is due to its wording.
		(B) The semantic information in each clause ($I_{M,i}$), as a function of clause number, for Charles Triplett's narrative ("Nice big blue .44" \cite{Labov2013}). 
		High variability is apparent. 
		Inset: histogram of the semantic information of each clause for all 20 Labov narratives, with the average (dashed line) close to 20 bits/clause.}
	\label{fig:sem_info}
\end{figure}

In Fig. \ref{fig:sem_info}A, we show the cumulative sums of $I_i$ for the 20 Labov narratives (for details about these narratives, see the Supplementary Information (SI), section \ref{si:narratives}). 
The average slope of these curves is approximately 40 bits/clause. 
Given that the average length of a clause in our dataset is close to 45 characters, we see that this results in information per letter of about 0.9 bits, which is compatible with Shannon's estimate (note that Shannon's values were obtained with 27 letter alphabet, a-z and space, while our narratives contain 72 characters, including capital letters, digits, and other characters.
Reducing the alphabet to 27 letters does not significantly change the result).

We now turn to the central point of our study, namely the information contained in the \textit{meaning} of a given clause ($M_i$) rather than in its exact phrasing ($C_i$). 
In general, the meaning of a particular clause is not unambiguous (e.g. saying 'I hate you' can mean anything from hate to love), but we assume that when the context of a clause is known, its meaning is uniquely determined ("single meaning approximation", see e.g. \cite{Kuhn2023,Niu2024}). 
We then need to estimate the conditional probability $P(M_i|c_{<i})$. 
Since there is no way to directly calculate this probability using LLMs (because they don't deal with meaning), one could decompose it into a sum of all possible phrasings that preserve the meaning of a given clause:
\begin{align}
	P(M_i|c_{<i}) = \sum_{C'_i} P(M_i|C'_i,c_{<i})P(C'_i|c_{<i}) \label{eq:m_cond_sum} = \sum_{\text{$C'_i$ means $M_i$}} P(C'_i|c_{<i})	\,\,, 
\end{align}
where the second equality is due to the "single meaning approximation" - $P(M_i| C'_i,c_{<i})=1$ if $ C'_i$ means $M_i$ (and $0$ otherwise).
In a recent paper \cite{Kuhn2023}, this equation was used together with a sampling-based approach (with only ten samples) for improved prediction of accuracy in the question-answering task.
In a more general setting, however, using the above definition is challenging as it requires the whole set of possible rephrasings of a given clause that preserves its meaning. 

Thus, we use a different approach by decomposing $P(C_i|c_{<i})$ as a sum over all possible meanings:
\begin{align}
	P(C_i|c_{<i}) &= \sum_{M'_i} P(M'_i|c_{<i})P(C_i|M'_i,c_{<i})  \label{eq:c_cond_sum} = P(M_i|c_{<i})P(C_i|M_i,c_{<i})  \,\,.
\end{align}
where the second equation follows again from the single meaning approximation, so that $P(C_i| M'_i,c_{<i})\ne 0$ only for $ M'_i=M_i$.
We can interpret Eq. \eqref{eq:c_cond_sum} intuitively: when a speaker composes a narrative, first they come up with an idea for the next clause ($M_i$) and then decide on its exact wording ($C_i$). 
The implicit assumption that we make here is that there are potentially multiple different wordings that communicate the same (or similar enough) meaning, otherwise the above equation would be simply reduced to $P(C_i|c_{<i}) = P(M_i|c_{<i})$. 

Using Eq. \eqref{eq:c_cond_sum}, we get the desired expression for semantic information of clause $i$, $I_{M,i}$:
\begin{align}
	I_{M,i} = -\log P(M_i|c_{<i}) = I_i - I_{W,i} \,\,, \label{eq:info_sem_clause}    
\end{align}
where $I_i = -\log P(C_i|C_{<i})$ is estimated directly by the token conditional probabilities from the LLM (see Eq. \eqref{eq:info_clause}), and $I_{W,i} = -\log P(C_i|M_i,c_{<i})$, the wording information of the clause, is calculated with $P(C_i|M_i,c_{<i})$.
This probability cannot be directly calculated with an LLM (again, as it does not handle meanings explicitly), however we designed the following procedure that allowed us to estimate it.

\begin{figure}[htb!]
	\centering
	\includegraphics[width=\textwidth]{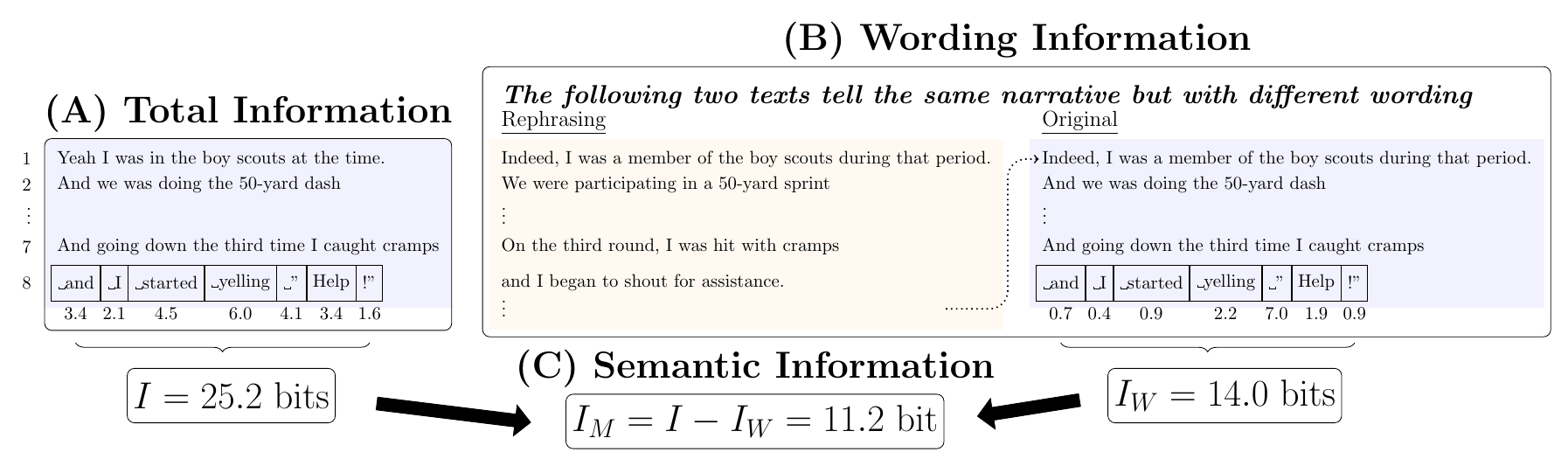}
	\caption{ {\it Estimating semantic information of a clause in a narrative.}
		We illustrate our method for estimating the semantic information of clause 8 for the "Boy scout" narrative \cite{Labov1966}.
		(A) Total information, $I$, is estimated by summing the information contents of each token in the clause, conditional on the preceding clauses.
		(B) Wording information, $I_W$, is estimated by summing the information contents of each token in the clause, conditional on a context which is supposed to constrain the meaning of the clause.
		The context includes a rephrased version of the narrative, the original preceding clauses, and a statement that the rephrasing conveys a similar meaning to the original.
		(C) Semantic information, $I_M$, is the difference between the total and wording information, see Eq. \eqref{eq:info_sem_clause}.
	}
	\label{fig:method}
\end{figure}

First, we generated a rephrasing of a narrative that closely preserves its meaning.
For this purpose, we used ChatGPT (gpt-4) \cite{GPT4_2023}, gave it the narrative segmented into clauses, and asked for a rephrased segmented narrative where the corresponding clauses retain the same meaning (the prompt is provided in the SI, section \ref{si:sem_info}). 
After obtaining the rephrased version of the narrative (N'), we went back to Llama and presented the following prompt to it: "The following two texts tell the same narrative but with different wording: N'… N" (precise prompt is given in the SI, section \ref{si:sem_info}).
Finally, given the above context, we computed the conditional probabilities for each clause in N. 
The rationale for this approach is that in the context of this prompt, the meaning of each clause of the narrative N is fixed by the previous context, and hence, the conditional probability for this clause should be equal to $P(C_i|M_i,c_{<i})$ (see Fig. \ref{fig:method} for the illustration of our method.)

In Fig. \ref{fig:sem_info}A, we show the cumulative sums of the semantic information $I_{M,i}$, for the 20 Labov narratives, with the average slope 20 bits/clause. 
One can see that roughly half of the information contained in each clause is due to its meaning, while the rest is the wording part.
The average semantic information rate ranges between 10 and 22 bits/clause.
There is, however, a much wider range of semantic information among individual clauses within each narrative, as shown in  Fig. \ref{fig:sem_info}B.
In Fig. \ref{fig:sem_info}B, we show an example for one narrative, whereas, in the inset, we plot the histogram for all the clauses in the 20 Labov narratives\footnote{
	Of the 1132 clauses, 4 were larger than 80 bits/clause, two slightly larger than 80, and two at about 100 bits/clause, and are not shown in the histogram.}.
In the following section, we present some interesting observations concerning this variability between clauses.

\section{Predictable and surprising clauses}
\begin{figure}[htb!]
	\centering
	\includegraphics[width=\textwidth]{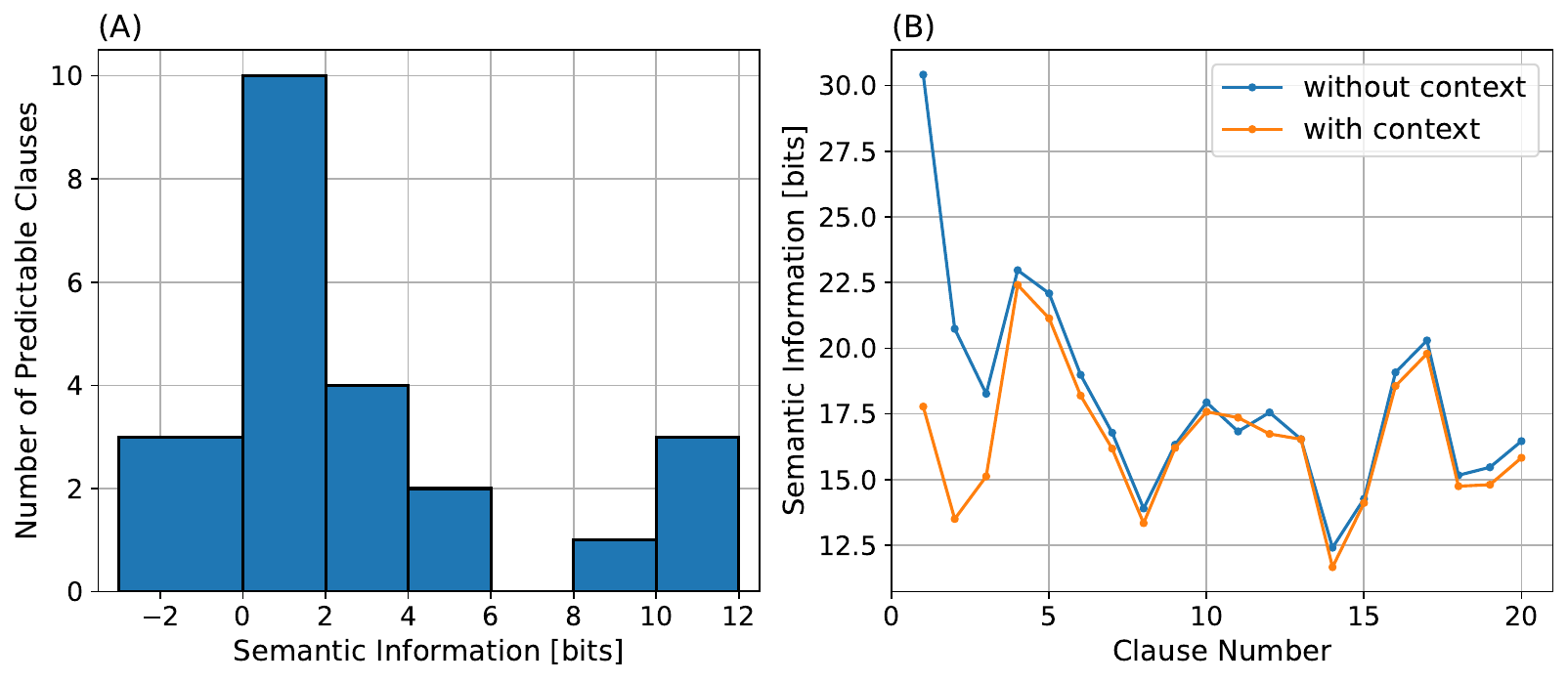}
	\caption{ {\it Predictable and surprising clauses in Labov narratives}. 
		(A) Semantic information ($I_M$) histogram of predictable clauses, which are clauses whose meaning can be guessed from the preceding part of the narrative with high probability (see text). 
		Predictable clauses have mostly semantic information close to zero.
		(B) Semantic information as a function of clause number, averaged over the 20 Labov narratives. 
		On average, semantic information is highest for the first clause and has a downward trend as the narrative unfolds (blue curve).
		However, with initial context as a guiding question initiating the narrative, this trend disappears.}
	\label{fig:predict_surprise}
\end{figure}

Curiously, some of the clauses contain very little information, see Fig. \ref{fig:sem_info}B, indicating that the meaning of these clauses is strongly constrained by context and could, therefore, be reasonably guessed. 
As a way of validating our method, we randomly sampled 543 clauses with different values of $I_M$ and asked ChatGPT to guess their meaning from the preceding part of the narrative (the prompt is given in the SI, section \ref{si:predictable}). 
ChatGPT correctly predicted the meaning of 31 clauses on the first attempt. 
Some of them appeared to be impossible to guess for the authors, which, we reasoned, might result from narratives potentially appearing in the training set of the LLM. 
We therefore performed a human guessing experiment on these clauses and found that they could predict the meaning of 23 of them with a single try (see the SI, section \ref{si:predictable}, for elaboration on our methodology). 
Semantic information of these clauses is shown in Fig. \ref{fig:predict_surprise}A. 
One can see that most of the predictable clauses indeed have small values of $I_M$.
Note however that there are three predictable clauses with semantic information between 10 and 12 bits, which indicates that our method sometimes results in wrong estimates of information.
The same can be said about clauses with negative semantic information (see Fig. \ref{fig:sem_info}A).

On the opposite side of the spectrum, clauses containing the highest information should be the most difficult to guess.
We cannot check this conjecture directly, however, because a vast majority of clauses cannot be guessed by humans in any reasonable number of attempts; in other words, we cannot directly apply Shannon's experimental method to clauses.
One could assume, however, that the beginning of a narrative should overall be more informative than the later parts, at least if the general context of the narrative is not known in advance. 
Indeed, we found that, on average, the first clause in the narrative has the highest information, with information plateauing after around ten clauses, as shown by the blue curve of Fig. \ref{fig:predict_surprise}B.
However, when we include initial context as the guiding question that initiated the narrative, the first clauses are no longer more informative on average, as shown by the orange curve (for elaboration, see SI, section \ref{si:initial_context}).

In our approach presented above, we used a rephrasing of the whole narrative to constrain the meaning of clause $C_i$, but we found that one can use only a partial rephrasing up to (and including) clause $C_i$ and get similar results, as detailed in the SI, section \ref{si:partial}.
Moreover, in SI, section \ref{si:consistency}, we showed that using another rephrasing of a narrative, obtained with the same prompt on the rephrased version, also results in similar values of semantic information. 

All the results presented above were obtained with Llama-3.1-70B-Instruct language model, which is a model post-trained for chat (chosen for its accessibility compared to the pre-trained version). 
In the Supplementary Information, section \ref{si:other_llms}, we show that using other LLMs, both pre- and post-trained, give similar results.

\section{Summary and future directions}
We developed a technique for estimating the information communicated by the meaning of clauses while ignoring their precise phrasing.
We believe this better describes human communication than the total information contained in each clause. 
We applied this procedure to a set of spoken narratives and found that, on average, meaningful information is about $20$ bits for a clause, or about one-half of the total.
We also found that different clauses contain vastly different semantic information, from $0$ to $100$ bits per clause. 
In particular, clauses that can be guessed by human subjects based on the previous context usually communicate a small amount of semantic information. 
We also found that clauses at the beginning of a narrative, without initial context, on average, convey more semantic information than subsequent clauses.

It will be interesting to apply our technique to other forms of verbal communication, such as dialogues.
With some modifications, one could extend this approach to quantitatively evaluating summaries, which could be used for the development of summarization programs.

\section*{Acknowledgements} 
We thank Drs. Mikhail Katkov and Tankut Can for helpful comments. 
M.T. is supported by the Simons Foundation, MBZUAI-WIS Joint Program for Artificial Intelligence Research and Foundation Adelis.

\bibliographystyle{unsrt}  
\bibliography{references}  

\clearpage
\appendix

\setcounter{section}{19}

\section*{Supplementary Information}
\setcounter{figure}{0}
\counterwithin{figure}{section}
\setcounter{table}{0}
\counterwithin{table}{section}

\etocsettocdepth{2}
\localtableofcontents  

\subsection{Labov narratives} \label{si:narratives}
We use 20 Labov stories: narrative number 6 in the paper \cite{Labov1966}, and 19 narratives from chapters 3-9 in the book \cite{Labov2013} (the 'narrative' of Seal: "In vain I tried to tell you" from chapter 4 was omitted as it is a traditional fictional tale and not a personal experience narrative).
These are personal experience narratives that describe events of danger of death, escalation of violence, sudden confrontation with death, premonitions and communication with the dead, and broader community conflicts.

The narratives were told during an interview, after the interviewer asked a leading question such as a danger of death question ("Were you ever in a situation where you thought you were in serious danger of being killed?").
Therefore, they are spontaneously spoken in a non-formal, conversational manner, with the local vernacular of the speaker, containing non-standard and sometimes outdated phrases. 
Moreover, they were accurately transcribed by Labov, and thus, the text contains repetitions, disfluencies, and self-corrections, as well as non-verbal sounds (like laughter) and verbal or non-verbal feedback from the interviewer.

Consequently, the narratives were slightly preprocessed: we removed non-verbal sounds and non-informative feedback (from the interviewer or other present people, e.g. "Mhm", "Wow!", but also informative feedback which the narrator then repeats).
We also carried the following edits:
\begin{itemize}[leftmargin=*, topsep=0pt]
	\setlength\itemsep{0em}	
	\item Gerry Furlow's narrative ("Injuries that could not be overcome "): the [gesture] transcriptions were removed as they don't add much information, while "he got me across here [stomach]" in clause 30 was changed to "he got me across the stomach".
	\item Melinda Dalphonso's narrative ("The falling out "): it is stopped in the middle of clause 18, as the rest is mainly the narrator's sister retelling with the narrator the main events.
	\item Margie Knott's narrative ("Her confrontation with the neighbors "): in clauses 43 and 96, "we were stuck into it" was replaced by "we were fightin'", and clauses 80 and 140 were minimally edited to include the feedback from the narrator's husband.
	\item Gloria Stein's narrative ("They stoned the house "): the transcription note [break in the recording] was removed as the narrative continues coherently afterwards.
	\item Rose Norman's narrative ("The death of her younger sister" ): the beginning of clause 38 is followed by an interviewer's question and her long (and repetitive) answer and thus it was shortened as if she explains shortly without being questioned. 
	Also, the quotes in Yiddish were replaced by their given English translations.
	\item Mary Costa's narrative ("The death of her youngest daughter "): in clause 47, "Oh and [breath] had uh everything in a bag", [breath] was replaced by "I". 
	In clause 127, "Her, Rita [shows picture]" was replaced by "Her, Rita, in this picture". 
	In clause 174, the transcription [Discussion of brother-in-law's family] was removed.
\end{itemize}

Labov segmented the narratives so that each independent clause and all subordinate clauses dependent on it are grouped under the same number \cite{Labov2013}.
In this paper, we refer to each such numbered text as a "clause", although sometimes it will contain multiple clauses (one independent clause and at least one dependent clause).
Below are a few "clauses" from Labov segmentation of the narrative of Gloria Stein \cite{Labov2013}, where independent and dependent clauses are marked by [\textit{ind.}] and [\textit{dep.}], respectively.

\begin{example}[ex:clauses]{Independent and dependent clauses}
	Gloria Stein's narrative: "They stoned the house". \\
	\begin{tabular}{l l}
		1. & We moved down to Third and uh Wharton. [\textit{ind.}] \\
		\dots & \\
		13. & And then of course they – they – well they stoned the house [\textit{ind.}] \\
		14. & And I had sense enough to get the kids into the stairway [\textit{ind.}] \\
		& so that they wouldn't get hurt by flyin' gla- glass. [\textit{dep.}] \\
		\dots & \\
		24. & but I remember saying a prayer [\textit{ind.}] \\
		& as I went out the front door [\textit{dep.}] \\
		& because I didn't know [\textit{dep.}] \\
		& what was gonna hit me [\textit{dep.}] \\
		& when I went out that door. [\textit{dep.}] \\
		\dots & \\
		40. & And then when he understood what the situation was [\textit{dep.}] \\
		& he turned to my mother [\textit{ind.}] \\
		\dots & \\
	\end{tabular}
\end{example}

\subsection{Estimation of semantic information} \label{si:sem_info}
The semantic information $I_M$ was estimated using the following procedure. We generated a narrative rephrasing using ChatGPT, and then calculated the conditional probabilities of each clause in the "wording information prompt" that includes the rephrased narrative followed by the original, obtaining the wording information $I_{W,i}$ for each clause $C_i$. We estimate the total information, $I_i$, by the conditional probabilities of each clause $C_i$ in the context of the previous clauses in the narrative. Finally, the semantic information was estimated from Eq. \eqref{eq:info_sem_clause}: $I_{M,i}=I_i - I_{W,i}$. Below, we give a detailed account of this procedure.

\subsubsection{Conditional probabilities} \label{sec:cond_probs}
\newcommand{\tbox}[2]{\overset{#1}{\boxed{#2}}}
\newcommand{\ttbox}[2]{\overset{\text{#1}}{\boxed{\strut\text{#2}}}}

First, we briefly overview how conditional probabilities are calculated in auto-regressive (decoder-only) LLMs, see e.g. \cite{Radford2019,Llama3_2024}. Given some text $T$, the LLM first segments its input text into tokens (tokenization), $T=\boxed{x_0} \boxed{x_1} \cdots \boxed{x_N}$,
where usually tokens correspond to words, and the first token ($x_0$) is a begin-of-text token. Then, in a single forward pass, it calculates the conditional log probability of each token (except the first) given the preceding tokens, $\log P(x_i|x_{<i})$. 
In this paper, we use the information content of each token (negative log-likelihood), $I_{x,i}=-\log P(x_i|x_{<i})$, and represent the text as $T=\tbox{I_{x,1}}{x_1} \tbox{I_{x,2}}{x_2} \cdots \tbox{I_{x,N}}{x_N}$ (removing the first begin-of-text token $x_0$, as its probability is not calculated).

For example, the information contents of the tokens (in bits per token) composing the first clause in the narrative of Doug D'Almaggio \cite{Labov2013} are given by:
\begin{align*}
	\ttbox{9.9}{So} \ttbox{6.7}{\textvisiblespace I} \ttbox{3.1}{\textvisiblespace was} \ttbox{3.6}{\textvisiblespace at} \ttbox{0.8}{\textvisiblespace the} \ttbox{7.8}{\textvisiblespace bar} \ttbox{9.4}{.} \,\,,
\end{align*}
while the first clause in the narrative of Harold Shambaugh ("The Norwegian sailor") \cite{Labov2013}, which has a similar meaning, but is spoken with a local dialect's pronunciation, is given by:
\begin{align*}
	\ttbox{13}{Oh} \ttbox{9.6}{\textvisiblespace I} \ttbox{13}{\textvisiblespace w} \ttbox{16}{'s} \ttbox{13}{\textvisiblespace sett} \ttbox{0.3}{in} \ttbox{0.5}{'} \ttbox{5.1}{\textvisiblespace at} \ttbox{5.4}{\textvisiblespace a} \ttbox{1.8}{\textvisiblespace table} \ttbox{7.3}{\textvisiblespace drink} \ttbox{0.03}{in} \ttbox{0.2}{'} \,\,.
\end{align*}

The information values were calculated using Llama3.1-70B-Instruct, which is a state-of-the-art open-source LLM.
We use an open-source LLM to have full access to the conditional probabilities, which is not possible with proprietary LLMs.

Llama-3.1-70B-Instruct is a model fine-tuned (post-trained) for chat, which is more popular and hence accessible than the non-fine-tuned version.
A pre-trained (non-fine-tuned) LLM is trained only on the language modeling task, to predict the next token using different context sizes (different numbers of preceding tokens).
Simply put, the pre-training objective is to minimize the information content of the tokens ($I_{x}$) in natural language texts. 
A post-trained LLM is further trained on texts of chat between a user and an AI assistant, where the user instructs the assistant to fulfill various tasks in natural language.
In section \ref{si:other_llms}, we provide results of other LLMs which are non-fine-tuned.
We note that we use Llama-3.1-70B-Instruct without the chat prompt format, as if it were the non-fine-tuned version (the only special token we use is a begin-of-text token).

\subsubsection{Narrative rephrasing} \label{si:rephrasing}
In \ref{prompt:rephrase} we give ChatGPT (gpt-4) \cite{GPT4_2023} \code{narrative part} and ask it to generate a paraphrase. \code{narrative part} is the original narrative numbered using Labov segmentation: "1. first clause 2. second clause ..." up to \code{N} clauses, and we ask that the rephrased narrative will also be numbered. We use zero temperature.

We found that this prompt works well for short narratives (less than 50 clauses), but for longer narratives, the quality of the rephrasing is compromised.
Also, sometimes, the rephrased version contains a different number of clauses from the original.
Therefore, we segment long narratives into $\lceil L/L_c \rceil$ equal parts, where $L_c=50$, and use \ref{prompt:rephrase} for each part separately. To assist with the rephrasing, when the part is in the middle of the narrative, we supply a summarization of the narrative up to that point, \code{summary} (otherwise, we remove the summary line from the prompt). The summarization was also generated by ChatGPT (gpt-4) via a different prompt.

\begin{prompt}[prompt:rephrase]{Narrative Rephrasing}
	\textbf{\textit{system:}} You are helping in scientific analysis, so please be precise. \\
	\textbf{\textit{user:}} You will be given a part of a narrative, segmented into linguistic clauses and numbered from 1 to \code{N}. \\
	Your task is to generate a paraphrase of this part of the narrative, using different wording (lexical diversity) and phrasing (syntactic diversity), but keeping the meaning essentially the same. \\
	You should keep the numbering of the clauses in the paraphrase. \\
	When the part is in the middle of the narrative, you will be given a summary of the narrative up to that point. \\
	Summarized Narrative so far: '''\code{summary}''' \\
	Part to paraphrase: '''\code{narrative part}'''
\end{prompt}

We checked that each rephrased narrative contained the same number of clauses as the original. Only in one case (narrative of Doug D'Almaggio, $L=31$ clauses \cite{Labov2013}) it did not, so we repeated the process with $L_c=25$ (instead of $L_c=50$).

We also manually went over a sample of the generated rephrased clauses to see that they made sense.
Below, we give a side-by-side comparison of the original and rephrased narratives for two examples: a short narrative, "Boy scout" \cite{Labov1966}, and a long narrative, Gloria Stein: "They stoned the house" \cite{Labov2013}.
Generally, the rephrasing is of good quality.
It conveys essentially the same meaning as the original and uses different wording.
In the "Boy scout" we notice a slight shortcoming in conserving the meaning: while the original "Boy scout" narrative is about a swimming race, ChatGPT misinterpreted it as a running race, making clause 13 a narrative inconsistency.
Interestingly, in the recall experiment of this narrative \cite{Georgiou2023}, many subjects also had the same misinterpretation (maybe because the words used throughout the original narrative, such as dash and racing, are better associated with running and not swimming), and then usually did not recall clause 13. 

Even though we asked in the prompt to include syntactic diversity, the paraphrases are mostly lexically different, while differences in syntax are less frequent.
Moreover, in the prompt, we didn't specifically ask to conserve the narrative's style; indeed, we noticed a change in style. 
The rephrased version is formal and edited (e.g., omits the disfluencies or self-corrections) as opposed to the original, which is casual and spontaneously spoken.
We leave improving the quality of the rephrasings and analyzing the sensitivity of semantic information to lexical, syntactic, and style changes for future studies.

\begin{example}[ex:boyscout]{Boy scout}
	\begin{tabular}{p{0.48\textwidth} p{0.48\textwidth}}
		\multicolumn{1}{c}{\textbf{Original}} & \multicolumn{1}{c}{\textbf{Rephrased}} \\
		1. Yeah I was in the boy scouts at the time. & 1. Indeed, I was a member of the boy scouts during that period. \\
		2. And we was doing the 50-yard dash & 2. We were participating in a 50-yard sprint \\
		3. racing & 3. competing \\
		4. but we was at the pier marked off & 4. at a designated pier. \\
		5. and so we was doing the 50-yard dash. & 5. Hence, we were engaged in the 50-yard sprint. \\
		6. There was about 8 or 9 of us you know, going down, coming back. & 6. There were approximately 8 or 9 of us, running forth and back. \\
		7. And going down the third time I caught cramps & 	7. On the third round, I was hit with cramps \\
		8. and I started yelling "Help!" & 8. and I began to shout for assistance. \\
\end{tabular}
\tcbbreak
\begin{tabular}{p{0.48\textwidth} p{0.48\textwidth}}
	9. but the fellows didn't believe me you know. & 9. However, my peers didn't take me seriously. \\
	10. They thought I was just trying to catch up because I was going on or slowing down. & 10. They assumed I was merely trying to keep pace because I was either advancing or slowing down. \\	
	11. So all of them kept going. & 11. Consequently, they all continued running. \\
	12. They leave me. & 12. They abandoned me.  \\
	13. And so I started going down. & 13. As a result, I began to sink. \\
	14. Scoutmaster was up there. & 14. The scoutmaster was present. \\
	15. He was watching me. & 15. He was observing me. \\
	16. But he didn't pay me no attention either. & 16. But he also disregarded me. \\
	17. And for no reason at all there was another guy who had just walked up that minute... & 17. Unexpectedly, another individual who had just arrived at that moment... \\
	18. He just jumped over & 18. He spontaneously leaped over \\
	19. and grabbed me. & 19. and rescued me.
	\end{tabular}
\end{example}

\begin{example}[ex:stein]{Gloria Stein: "They stoned the house"}
	\begin{tabular}{p{0.48\textwidth} p{0.48\textwidth}}
		\multicolumn{1}{c}{\textbf{Original}} & \multicolumn{1}{c}{\textbf{Rephrased}} \\
		1. We moved down to Third and uh Wharton. & 1. We relocated to the intersection of Third and Wharton. \\
		2. We had gotten moved & 2. Our move had been completed,  \\
		3. but it was like – you remember I told you my father drank & 3. but it was during the day - recall I mentioned my father's drinking habit. \\
		4. but it was still day & 4. Despite it being daytime, \\
		5. and after he had gotten in and put the furniture where it was supposed to go my mother left, to go to market to get some food to bring into the house. & 5. after my father had settled in and arranged the furniture, my mother departed to the market to purchase groceries for our new home. \\
		6. And my father went to take the horses – take the team of horses back. & 6. Meanwhile, my father left to return the horses - the pair that had helped us move. \\
		\dots & \dots \\
	\end{tabular}
	\small \lbrack After the mother left for the market, the house was stoned. Gloria ran to her, and they met a policeman who warned them about the neighborhood. The mother refused to move and decided to go to the police station.\rbrack \normalsize \\
	\begin{tabular}{p{0.48\textwidth} p{0.48\textwidth}}
		\dots & \dots \\
		66. And we went – she took me and we went to the police station. & 	66. She took me along to the police station. \\
		67. And they also had police wagons with horses attached to them. Believe it or not in those days. & 67. They even had police wagons drawn by horses in those days. \\
		68. So my mother said to the captain, "What are you going to do about it?" & 68. My mother asked the captain, "What action will you take?" \\
		69. And he says "Well lady there's nothing we can do about it." & 69. He replied, "Unfortunately, there's nothing we can do." \\
		70. And she says "You mean to tell me that I have to live under those circumstances and there's nothing that you can do about it?" & 70. She retorted, "Are you implying that I must endure these conditions and you're incapable of providing any assistance?" \\
	\end{tabular}
	\begin{tabular}{p{0.48\textwidth} p{0.48\textwidth}}
		71. She says – he says "Well uh uh the only thing I can tell you to do is to move." & 71. He suggested, "The only advice I can offer is to relocate." \\
		\dots & \dots \\
	\end{tabular}
	\small \lbrack The mother refused and insisted on a police escort home. They arrived in front of curious neighbors, and the mother declared they were staying.\rbrack \normalsize \\
	\begin{tabular}{p{0.48\textwidth} p{0.48\textwidth}}
		\dots & \dots \\
		125. So in the meantime – not in the meantime but during during the period of time that we were there, my father became sick. & 125. During our time there, my father fell ill. \\
		126. And the neighbors were as nice as anybody could ever have been. & 126. The neighbors turned out to be as kind as anyone could hope for. \\
		127. When there was trouble they responded just like anyone else. & 127. They reacted to trouble just like anyone else would. \\
	\end{tabular}
	\tcbbreak
	\begin{tabular}{p{0.48\textwidth} p{0.48\textwidth}}
		128. But that was our welcoming to the neighborhood. & 128. But that was our introduction to the neighborhood. \\
		129. And it was terrible & 129. It was a terrible experience, \\
		130. and it certainly was frightening for me. & 130. and it was certainly scary for me. \\
	\end{tabular}
\end{example}

\subsubsection{wording and semantic information}
We use \ref{prompt:wording} to estimate the wording information of the clauses in the narrative, where \code{rephrased} is the rephrased narrative and \code{original} is the original one.

\begin{prompt}[prompt:wording]{Wording information}
	The following two texts, separated by \text{-}\text{-}\text{-}, tell the same narrative but with different wording. \\
	\code{rephrased} \\
	\text{-}\text{-}\text{-} \\
	\code{original}
\end{prompt}

The wording information of clause $C_i$, $I_{W,i}$, is estimated via Llama3.1-70B \cite{Llama3_2024}, by summing up the information contents of all the tokens corresponding to clause $C_i$ that appear in \code{original} in the context of \ref{prompt:wording}: 
\begin{align*}
	I_{W,i}=\sum_{\substack{\text{token $x_j$ of clause $C_i$} \\ \text{in \code{original}}}} I_{x,j} \,\,,
\end{align*}
where the token information content is $I_{x,j}=-\log P(x_j|x_{<j})$, as discussed in section \ref{sec:cond_probs}. 
We call it the wording information, because \code{rephrased} is supposed to constrain the meaning of each clause in \code{original} in \ref{prompt:wording}, so that the conditional probability for clause $C_i$ in \code{original} should be equal to $P(C_i|M_i,c_{<i})$, which is the probability to use the exact original wording, given the knowledge of the meaning of the clause (and the preceding context). 

On the other hand, the total information, $I_i$ is estimated with Llama3.1-70B \cite{Llama3_2024}, by summing up the information contents of all the tokens corresponding to clause $C_i$ that appear in the original narrative alone (i.e. without the preceding meaning constraining context). Then, the semantic information, $I_{M,i}$ is estimated from Eq. \eqref{eq:info_sem_clause}:  $I_{M,i}=I_i - I_{W,i}$.

To illustrate the process, we give an example of this calculation for clause 7 in the "Boy scout" narrative \cite{Labov1966}: "And going down the third time I caught cramps". Using the original and rephrased narrative given in \ref{ex:boyscout}, we construct \ref{prompt:wording}, and calculate via Llama the information contents of all the tokens corresponding to clause 7 that appear in \code{original} in the context of \ref{prompt:wording}:
\begin{align*}
	\ttbox{1.9}{\textvisiblespace And} \ttbox{11.2}{\textvisiblespace going} \ttbox{1.6}{\textvisiblespace down} \ttbox{1.0}{\textvisiblespace the} \ttbox{0.1}{\textvisiblespace third} \ttbox{0.4}{\textvisiblespace time} \ttbox{0.9}{\textvisiblespace I} \ttbox{5.8}{\textvisiblespace caught} \ttbox{1.5}{\textvisiblespace cr} \ttbox{0.1}{amps} \,\,,
\end{align*}
Summing up, we get $I_{W,7}=24.4$ bits. Then, we calculate with Llama the information contents of all the tokens corresponding to clause 7 in the original narrative alone, with only the previous clauses as context:
\begin{align*}
	\ttbox{1.8}{\textvisiblespace And} \ttbox{11.1}{\textvisiblespace going} \ttbox{0.8}{\textvisiblespace down} \ttbox{3.7}{\textvisiblespace the} \ttbox{4.3}{\textvisiblespace third} \ttbox{0.7}{\textvisiblespace time} \ttbox{2.2}{\textvisiblespace I} \ttbox{7.4}{\textvisiblespace caught} \ttbox{5.4}{\textvisiblespace cr} \ttbox{0.9}{amps} \,\,.
\end{align*}
Thus, $I_{7}=38.3$ bits and finally $I_{M,7}=I_7-I_{W,7}=13.9$ bits.

\subsection{Identification of predictable clauses} \label{si:predictable}
The 20 Labov narratives contain 1132 clauses.
Instead of analyzing all of them, we sampled the clauses according to their semantic information. 
We sampled 40 clauses per bin, for 14 bins with the following bin boundaries: 0, 2, 4, 6, 8, 10, 12, 14, 16, 20, 25, 30, 40, 60, and 200 bits.
The bins 0-2, 2-4, and 60-200 bits have 33, 39, and 17 clauses, respectively, so we analyzed all of them. 
There were also 14 clauses with negative semantic information (between -3 and 0 bits). 
Therefore, we define another bin for these clauses, with a total of 15 bins and 543 clauses.
Eight of the sampled clauses were the first ones in the narrative and thus were omitted, as it is impossible to predict them without context.

\subsubsection{ChatGPT prediction}
We then used ChatGPT (gpt-4o, zero temperature) \cite{GPT4_2023} to judge whether these clauses are predictable with the following procedure.
First, we used \ref{prompt:continuation} to obtain the most plausible continuation for the next clause given \code{narrative part}, which is the part paused just before the original clause under consideration, divided into numbered clauses.

\begin{prompt}[prompt:continuation]{Most plausible continuation}
	\textbf{\textit{system:}} [None given]\\
	\textbf{\textit{user:}} In this task, you are presented with a narrative divided into numbered clauses. \\
	The narrative is paused at a certain point, and your task is to generate the most plausible continuation for the next clause only. \\
	A clause should contain a single piece of information, a single action, etc.\\ \\
	Here is the narrative:"' \code{narrative part}'". \\ \\	
	Please output the continuation in the following format:
	The most plausible next clause is "'clause text'".
\end{prompt}

Second, we used \ref{prompt:same_meaning} to judge if \code{proposed clause}, the output of the previous prompt, conveys essentially the same meaning as \code{original clause}, giving as context \code{narrative part} (as was given in the previous prompt) to understand the meaning of the two continuations better.

\begin{prompt}[prompt:same_meaning]{Same meaning}
	Here is a narrative, divided into numbered clauses, and paused at a certain point:"' \code{narrative part}'". \\ \\
	Here are two possible continuations for the next clause: \\
	1. '''\code{original clause}''' \\
	2. '''\code{proposed clause}''' \\ \\
	Do they convey essentially the same meaning (wording/phrasing may differ)? \\
	Answer in a step-by-step manner. \\
	At the end of your answer provide a True/False decision in the following format: **Same meaning: True/False**.
\end{prompt}

In \ref{prompt:same_meaning} we used (zero-shot) chain-of-thought prompting \cite{Wei2022} to increase the likelihood of correct judgments, as we detected clear errors in a prompt which instructs for a short True/False answer.
Out of the 535 clauses, ChatGPT judges 45 of them to be predictable using the above prompt.
We verified the same meaning judgment by going over these 45 clauses.
We detected 14 errors (false positives) and 31 clauses in which ChatGPT could predict the meaning of the next clause in a single attempt. 
Below are a few examples out of these 31 clauses from the same narrative of CI Sr.: "The notice in the New York Times" \cite{Labov2013}:

\begin{example}[ex:predictable]{Predictable clauses in C.I.}
	1. Yeah there is an instance.\\ 
	2. Dad was being driven out from town – ah – by his chauffeur.\\ 
	3. This was a good many years ago.\\ 
	4. And he had the New York Times\\ 
	5. and he read in the New York Times\\ 
	6. and noticed that – the death of a person whom he knew but he knew was a very close friend of George Jensen.\\ 
	7. George Jensen lived in Chestnut Hill.\\ 
	8. So Dad said to the chauffeur "Stop at Mr. Jensen's house on the way home 'cause I want to commiserate with him." \\ 
	9. So they did stop\\ 
	\textbf{9'. ChatGPT: "The chauffeur nodded and changed the route accordingly." (semantic information of 5.4 bits)}\\
	10. and Dad went in\\ 
	11. and Dad said "George, I'm so sorry to hear about the death of —" \\ 
	12. I don't know his name\\ 
	13. and George Jensen said "I don't know what you're talking about.\\ 
	14. If he had died I would have been one of the first people to know." \\ 
	15. And Dad said "Well it's in the newspaper,\\ 
	16. I'll go out to the car\\ 
	17. and get the newspaper." \\ 
	18. Went out\\ 
	19. and got the newspaper\\ 
	\textbf{19'. ChatGPT: "and retrieved the newspaper." (semantic information of 1.8 bits)}\\
	20. came back\\ 
	\textbf{20'. ChatGPT: "and brought it back inside." (semantic information of 3.4 bits)}\\
	21. and he and George went through the newspaper.\\ 
	22. No sign of this death notice.\\ 
	23. And just as they were finished perusing it the telephone rang from somebody in New York telling George Jensen that, guess he'd died.\\ 
	24. But there was nothing in the newspaper.\\ 
	25. Dad brought the newspaper home.\\ 
	26. My sister – guess George was home at the time – and I all went through the newspaper meticulously.\\ 
	27. Couldn't find anything. \\
	\textbf{27'. ChatGPT: "Still, there was no mention of the death notice." (semantic information of 0.74 bits)}
\end{example}

As discussed in the paper, a possible concern is that maybe the LLM had seen the narratives during training and thus is biased to output similar meaning continuations with higher probability.
Previous studies have demonstrated the verbatim memorization phenomenon in LLMs, which increases the more times the text is duplicated in the dataset \cite{Carlini2022}.
As memorization can cause copyright and privacy issues, LLMs try to eliminate it, with the most straightforward method (but only partially effective) deduplicating the training data \cite{Carlini2022}. 
Even with more sophisticated methods, verbatim memorization might be effectively eliminated, but not approximate memorization (memorization of text with a similar meaning to the original) \cite{Ippolito2022}.

\subsubsection{Human prediction}
Therefore, to further verify our results, we asked subjects to predict the next clause in a paused narrative, similar to the task presented to ChatGPT in \ref{prompt:continuation}. 
We recruited 13 participants online via Prolific (www.prolific.com), with the following screeners: primary language English, 20 or more previous submissions, and an approval rate above 99\% (around 100,000 active participants matched these criteria).
Each participant predicted 14 or 17 clauses from 6 different narratives, with a median time to complete of 40 minutes.
Each participant was compensated 8 USD for the task (median rate: 12 USD/hour).

The task was conducted online on the participants' browsers.
The participants were instructed: "This is a writing task. 
You will be presented with true personal experience narratives told by real people. 
Each narrative is segmented into clauses, which are simple sentences. 
The story is paused at a certain point, and your task is to predict the next clause. 
To be correct, you only need your predicted clause to convey essentially the same meaning as the one from the true story; wording or phrasing does not matter.".
These instructions were followed by remarks (in short: the narrative can contain slang, some numbered clauses also contain dependent clauses, take time before you answer, don't use AI assistants or any other trick), followed by a consent form to participate in this research.

Then, each participant was assigned to one of two groups.
The first group, with 7 participants, predicted 14 clauses from 6 narratives.
The second group, with 6 participants, predicted 17 clauses from 6 different narratives.
The participants predicted in total the 31 clauses that were found predictable, as described in the previous section.
For each prediction, the participant was given the previous numbered clauses in the narrative up to that point, and he entered his prediction for the next clause in a textbox.
When a narrative contained more than a single clause to predict, the already read clauses appeared in a gray background to remind the reader that he had read these clauses.

The responses of the participants were primarily usable.
The only issue was one participant from the first group, halfway through the task, started entering previous predictions (verbatim); these copied predictions were removed from the analysis.
Thus, 24 clauses have predictions from 6 participants, and 7 clauses have predictions from 7 participants.

For each participant's prediction, we manually judged whether it conveyed essentially the same meaning as the original clause.
We note two cases that required careful judgment.
The first is clause 14 in Knott: "So a' went over that night the same night they moved in," which was only partially predicted by four participants - e.g. "so i knock on the door and introduce myself" or "So we introduced ourselves and asked 'need anything from the shop?'" - and it was decided to be classified as not predicted.
The second is clause 81 in Norman: "So sure enough he went to Harry Pressger", where one prediction was correct, "My husband went to ask Harry.", and in two others, there was confusion between the narrator (the wife) and her husband - "So I went" and "So, my wife went to 803 South 2nd to meet Harry Pressger" - but otherwise they were correct, so it was decided to classify them as predicted.

There were 8 clauses that no participant could predict.
The 23 other clauses could be predicted usually by 2 to 4 participants (in detail, 1, 8, 6, 5, 2, 1 clauses were predicted by 1, 2, 3, 4, 5, 6 participants, respectively).
The fact that ChatGPT could predict these 8 clauses with a single attempt, but 6-7 participants could not, means that ChatGPT has better-than-average-human ability on this task, which can be caused by the memorization effect mentioned above.

\begin{figure}[htb!]
	\centering
	\includegraphics[width=0.5\textwidth]{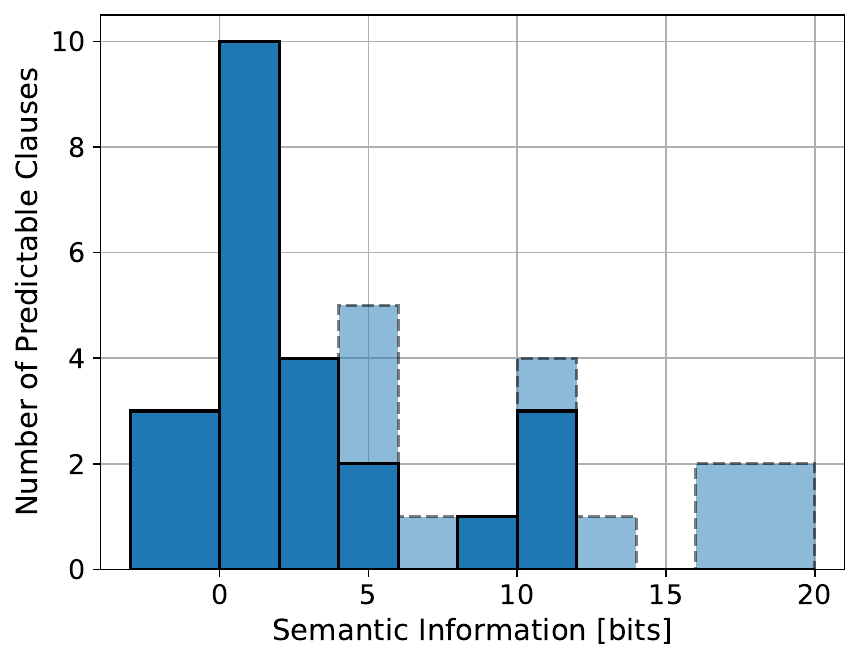}
	\caption{ {\it Predictable clauses by humans and ChatGPT}. 
		Semantic information ($I_M$) histogram of predictable clauses. 
		In blue are clauses correctly predicted by both ChatGPT and human participants, while in light blue are clauses predicted by ChatGPT but not by humans.}
	\label{fig:predictable}
\end{figure}

In Fig. \ref{fig:predictable}, we show a version of Fig. \ref{fig:predict_surprise}A  in which the clauses deemed predictable by ChatGPT but not by humans appear in light blue.
It can be seen that the 8 clauses that were not predicted by humans have systematically higher semantic information.

\subsection{Effect of initial context} \label{si:initial_context}
Most of Labov's narratives were told as a direct answer to the interviewer's question.
As previously mentioned, the most common question was the danger of death question: "Were you ever in a situation where you thought you were in serious danger of being killed?".
There were also questions about confrontation with death, premonitions, and relationships with neighbors, while few narratives were told voluntarily, without any guiding question \cite{Labov2013}.

The following table lists the different contexts with the various guiding questions we used for the different narratives. 
In the narratives of Ellen Laidlaw and Mary Costa, an initial answer from the interviewee is necessary to provide context before the narrative is told.
The narratives of Gloria Stein and Rose Norman do not appear in the table as they told their story unprompted \cite{Labov2013}.

\begin{tabular}{|p{0.71\textwidth} | p{0.25\textwidth}|}
	\hline
	\multicolumn{1}{|c|}{\textbf{Context}} & \multicolumn{1}{|c|}{\textbf{Narratives}}  \\ \hline
	Interviewer: Were you ever in a situation where you thought you were in serious danger of being killed? \newline Interviewee: & Schissel, Shambaugh, D{'}Almaggio, Furlow, Triplett, Tarentino, Boy-scout \\ \hline
	Interviewer: Were you ever in a situation where you thought \textit{someone was} in serious danger of being killed? \newline Interviewee: & Williams \\ \hline
\end{tabular}

\begin{tabular}{|p{0.71\textwidth} | p{0.25\textwidth}|}
	\hline
	Interviewer: Were you ever in a situation where you were suddenly faced with the fact of death? \newline Interviewee: & Hawkins, McCaffrey, Adamo \\ \hline
	Interviewer: Were you ever in a situation where you were suddenly faced with the fact of death? \newline Interviewee: \textit{No, but it happened to my mother.} & Laidlaw \\ \hline
	Interviewer: Is there anyone you know who gets a feeling that something is going to happen, and then it does happen? \newline Interviewee: & C.I., Dalphonso, Guyton, Hester \\ \hline
	Interviewer: Is your neighborhood friendly? \newline Interviewee:  Well no neighborhood's friendly, unless you assert yourself. & Knott \\ \hline
	Interviewer: You said your other daughter died. How long ago did she die? \newline Interviewee: & Costa \\ \hline
\end{tabular}

In Fig. \ref{fig:predict_surprise}B we showed that the initial context affects mainly the three first clauses, with the most effect on the first clause.
Adding the initial context erases the trend of the initial clauses being more informative.
The effect on the rest of the clauses (other than the first three) is low: the change in semantic information has an average of 0.2 bits and a standard deviation of 1.5 bits.

\subsection{Effect of partial rephrasing} \label{si:partial}
The method we proposed uses the complete rephrasing in \ref{prompt:wording} to estimate the wording information of each clause.
This method has the advantage of estimating the wording information of all the clauses in the narrative with a single prompt.
However, it might not be necessary to use the complete rephrasing to constrain the meaning of clause $C_i$ in the prompt.

To investigate it, we calculate the wording information of clause $C_i$ with a prompt similar to \ref{prompt:wording}, but \code{rephrased} is only a partial rephrasing up to (and including) clause $C_i$.
The wording information with a partial rephrasing deviates from the wording information with a complete rephrasing by $(-1.1\pm2.3)$ bits.
For the long stories (Knott, Stein, Norman, and Costa, with more than 65 clauses), we compared only the 30 first and 30 last clauses.
The deviations had no apparent trend with clause number or semantic information.

The above result means that providing rephrasings of the clauses after clause $C_i$ further constrains the meaning of clause $C_i$ on average, but only by a tiny amount.
The two methods result in very similar semantic information, and the method we used in the paper is preferred as it uses a single prompt instead of $L$ prompts for a narrative with $L$ clauses.

\subsection{Semantic consistency test} \label{si:consistency}
In a semantic consistency test of an LLM, we check if the result is not sensitive to a rephrasing of the prompt which preserves its meaning, see e.g. \cite{Jang2023}.
In our case, we want to test if our method of estimating semantic information is not sensitive to rephrasings of the narrative that preserve the meaning.

\begin{figure}[htb!]
	\centering
	\includegraphics[width=\textwidth]{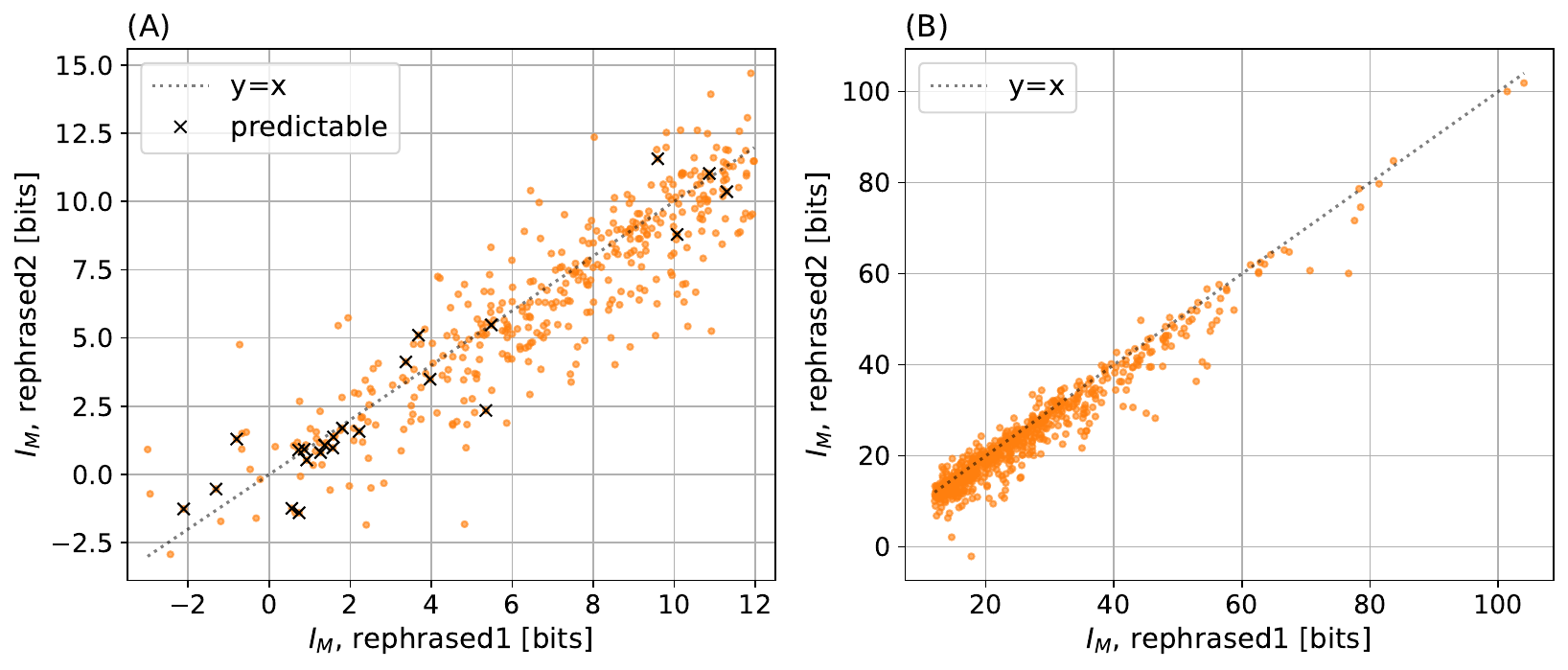}
	\caption{ {\it Semantic consistency test}.
		Scatter plot of semantic information ($I_M$) of the different clauses, estimated with another rephrasing (rephrased2), against semantic information estimated with the first rephrasing (rephrased1), in orange dots.
		The dotted gray line corresponds to an ideal consistency, $y=x$. 
		(A) Scatter plot between -3 and 12 bits, with the 23 predictable clauses found in section \ref{si:predictable} marked by X. 
		(B) Scatter plot for clauses with semantic information greater than 12 bits.
}
	\label{fig:consistency}
\end{figure}

We implemented a semantic consistency test by generating another rephrasing, rephrased2, and comparing the semantic information calculated with rephrased2 against the previous value. 
We generated rephrased2 using the same method outlined in section \ref{si:rephrasing}, but with the first rephrasing as the original narrative.
The results are shown in Fig. \ref{fig:consistency}, where we can see adequate agreement between the two rephrasings.
Quantitatively, for the 407 clauses with semantic information less than 12 bits, the semantic information deviates by $(-0.4\pm 1.6)$ bits (the deviation for the 23 predictable clauses is $(-0.2\pm 1.2)$ bits).
The relative deviation for the other 725 clauses is $(-7\pm13)$\%.

\subsection{Estimating with other LLMs} \label{si:other_llms}
\subsubsection{Comparison with a smaller model}
First, we compare semantic information estimated with Llama3.1-70B-Instruct to semantic information estimated with a smaller model, Llama3.1-8B-Instruct \cite{Llama3_2024}.
The results are shown in Fig. \ref{fig:llms}A and summarized in Table \ref{tab:llms}, case A.
We see that, on average, semantic information is very similar between the two models.
While there is some spread around this average, the spread of predictable clauses is smaller.
Hence, even when estimating with a smaller model, predictable clauses remain with mostly small values of semantic information.
Still, we notice two predictable clauses that changed from about 5 bits to around 10 bits, which may point to a limitation of the smaller model.

\subsubsection{Comparison with a non-fine-tuned model}
We compare semantic information estimated with the fine-tuned Llama3.1-8B-Instruct to semantic information estimated with a non-fine-tuned model, Llama3.1-8B \cite{Llama3_2024}.
We get a spread much like the previous case, with predictable clauses having an even smaller spread, as depicted in Fig. \ref{fig:llms}B and summarized in Table \ref{tab:llms}, case B.

\subsubsection{Comparison with a comparable Mistral model}
\begin{figure}[htb!]
	\centering
	\includegraphics[width=\textwidth]{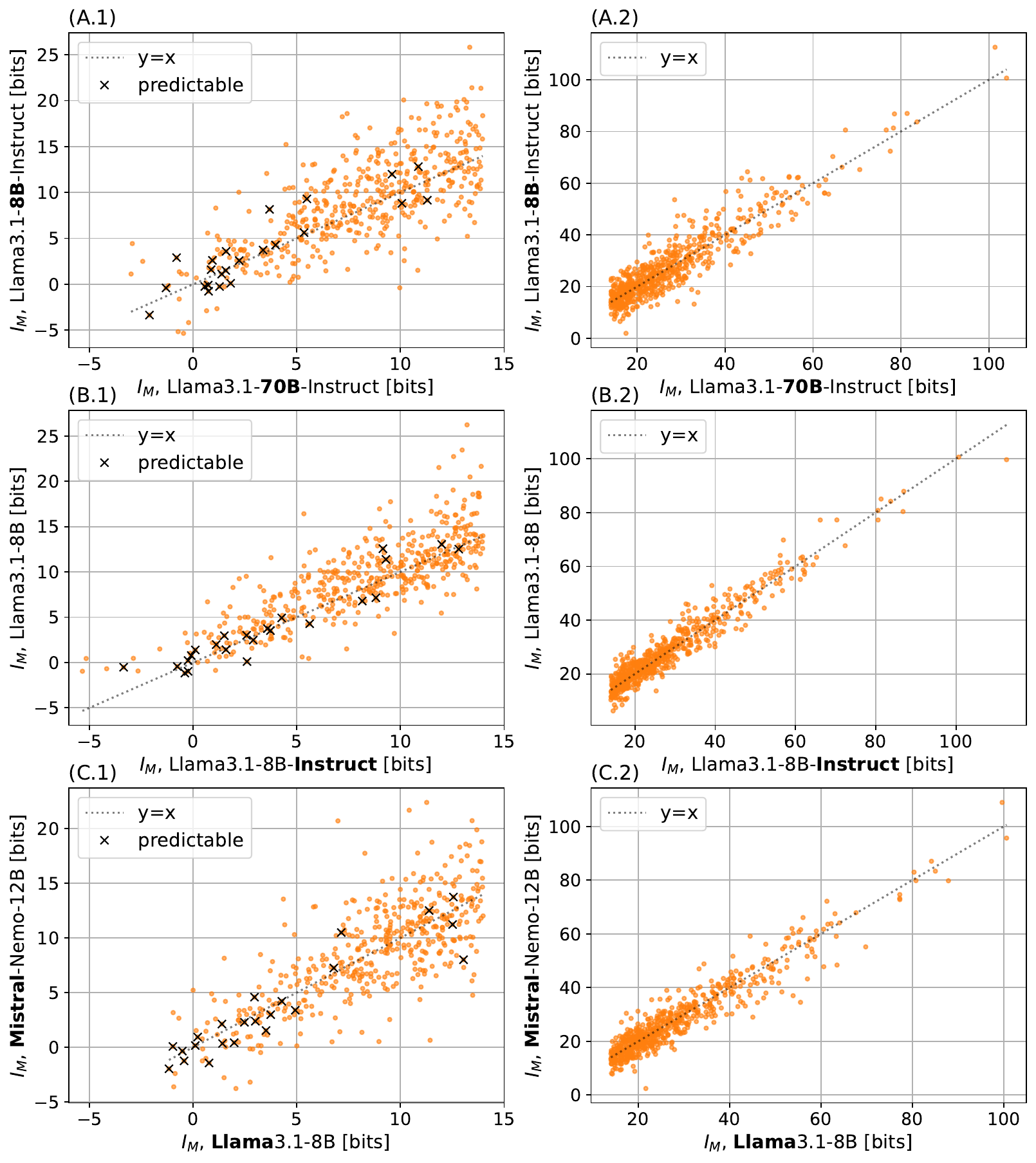}
	\caption{ {\it Estimating semantic information with other LLMs}.
		Scatter plot of semantic information ($I_M$) of the different clauses, comparing two different LLMs, in orange dots.
		The dotted gray line corresponds to equivalence, $y=x$.
		Three comparisons are shown:
		(A) Comparison between Llama3.1-8B-Instruct and  Llama3.1-70B-Instruct.
		(B) Comparison between Llama3.1-8B and Llama3.1-8B-Instruct.
		(C) Comparison between Mistral-Nemo-12B and Llama3.1-8B.
		For each comparison, we split the plot into two:
		(1) Scatter plot for clauses with semantic information less than 14 bits, with the 23 predictable clauses found in section \ref{si:predictable} marked by X.
		(2) Scatter plot for clauses with semantic information greater than 14 bits.
}
	\label{fig:llms}
\end{figure}

Finally, we also compare semantic information estimated with the Llama3.1-8B to semantic information estimated with an independently developed model, Mistral-Nemo-12B \cite{Nistral7B_2023,MistralNemo2024}, which is another state-of-the-art model with a size comparable to Llama3.1-8B.
We get spread similar to the previous cases, as shown in Fig. \ref{fig:llms}C and summarized in Table \ref{tab:llms}, case C.

\begin{table}[htb!]
	\centering
	\caption{ Comparison summary of the three comparisons in Fig. \ref{fig:llms}:
		(A) Comparison between Llama3.1-8B-Instruct and Llama3.1-70B-Instruct.
		(B) Comparison between Llama3.1-8B and Llama3.1-8B-Instruct.
		(C) Comparison between Mistral-Nemo-12B and Llama3.1-8B. \\
		We report mean$\pm$std of absolute semantic information deviations of (1) the 23 predictable clauses, (2) absolute deviations of clauses with $I_M<14$ bit (left of  Fig. \ref{fig:llms}), and (3) relative deviations of clauses with $I_M>14$ bit (right of Fig. \ref{fig:llms}).
	}
	\begin{tabular}{|c|c|c|c|}
		\hline
		\textbf{Case} & \textbf{Abs. Deviation [bit], predictable} & \textbf{Abs. Deviation [bit], $\mathbf{I_M<14}$bit} & \textbf{Rel. Deviation [\%], $\mathbf{I_M>14}$bit} \\ \hline
		(A) & $0.5\pm 1.9$ & $1.0\pm 3.3$ & $1\pm 21$ \\ \hline
		(B) & $0.3\pm 1.4$ & $1.0\pm 2.6$ & $1\pm 16$ \\ \hline
		(C) & $-0.3\pm 1.7$ & $0.1\pm 3.0$ & $1\pm 17$ \\ \hline
	\end{tabular}
	\label{tab:llms}
\end{table}

\clearpage

\clearpage
\subsection{Supplementary data}
Supplementary data is available at \href{https://github.com/DoronSivan/info\_narrative\_data}{github.com/DoronSivan/info\_narrative\_data}.
Data includes:
\begin{itemize}[leftmargin=*, topsep=0pt]
	\setlength\itemsep{0em}	
	\item The 20 Labov narratives with Labov segmentation into clauses ('original.csv').
	\item The two rephrasings that were generated as described in sections \ref{si:rephrasing} and \ref{si:consistency} ('rephrased1.csv' and 'rephrased2.csv', respectively).
	\item Total, wording, and semantic information for the various clauses ('semantic\_information.csv').
	The values are estimated using the method described in the paper, with Llama3.1-70B-Instruct and the rephrasing given in 'rephrased1.csv'.
	\item The ChatGPT (gpt-4o) predictions for the next clause for 535 sampled clauses ('gpt\_predictions.csv').
	Column 'correct\_gpt' is true when ChatGPT judges its prediction to convey essentially the same meaning as the original clause, and column 'correct\_gpt\_verified' is true when this judgment was also verified by the authors (see section \ref{si:predictable}).
	\item Human predictions for the next clause for the 31 clauses with 'correct\_gpt\_verified' equals true.
	The participant's predictions appear in the columns starting with 'p\_'.
	Column 'num\_answers' contains the number of participants that predicted this clause, and column 'num\_correct' is the number of participant's predictions the authors judge to convey essentially the same meaning as the original clause.
\end{itemize}
The files 'original.csv', 'rephrased1.csv' and 'rephrased2.csv' are put in a password-protected zip ('narratives.zip', password: labov-narratives) to make them less accessible as LLM training data.
\end{document}